\documentclass{midl} 

\usepackage{natbib}
\newcommand{\sig}[1]{\textbf{#1*}}

\usepackage{mwe} 
\usepackage{todonotes}
\usepackage{multirow}
\usepackage{booktabs}   
\usepackage{tikz}
\usetikzlibrary{backgrounds}

\usetikzlibrary{arrows.meta,positioning,fit}
\usepackage{graphicx}
\usepackage[justification=centering]{caption}

\title[Deep Learning Based CAC Detection: A Synthetic Approach]{Machine-Learning Based Detection of
Coronary Artery Calcification Using
Synthetic Chest X-Rays}

\midlauthor{\Name{Dylan Saeed\nametag{$^{1}$}}\Email{d.saeed@student.unsw.edu.au}\\
\Name{Ramtin Gharleghi\nametag{$^{3}$}} \Email{r.gharleghi@unsw.edu.au}\\
\Name{Susann Beier\nametag{$^{2}$}} \Email{s.beier@unsw.edu.au}\\
\Name{Sonit Singh\nametag{$^{1}$}} \Email{sonit.singh@unsw.edu.au}\\
\AND 
\addr $^{1}$ School of Computer Science and Engineering, University of New South Wales, Australia \\
\addr $^{2}$ School of Mechanical and Manufacturing Engineering, University of New South Wales, Australia \\
\addr $^{3}$ Independent Researcher, Sydney, Australia
}
\begin{document}
\maketitle
\vspace{-2em}
\begin{center}
\textbf{Preprint. Under review at MIDL 2026.}
\end{center}
\begin{abstract}
Coronary artery calcification (CAC) is a strong predictor of cardiovascular events, with CT-based Agatston scoring widely regarded as the clinical gold standard. However, CT is costly and impractical for large-scale screening, while chest X-rays (CXRs) are inexpensive but lack reliable ground truth labels, constraining deep learning development. Digitally reconstructed radiographs (DRRs) offer a scalable alternative by projecting CT volumes into CXR-like images while inheriting precise labels. In this work, we provide the first systematic evaluation of DRRs as a surrogate training domain for CAC detection. Using 667 CT scans from the COCA dataset, we generate synthetic DRRs and assess model capacity, super-resolution fidelity enhancement, preprocessing, and training strategies. Lightweight CNNs trained from scratch outperform large pretrained networks; pairing super-resolution with contrast enhancement yields significant gains; and curriculum learning stabilises training under weak supervision. Our best configuration achieves a mean AUC of 0.754, comparable to or exceeding prior CXR-based studies. These results establish DRRs as a scalable, label-rich foundation for CAC detection, while laying the foundation for future transfer learning and domain adaptation to real CXRs. 

\end{abstract}

\begin{keywords}
Coronary artery calcification, digitally reconstructed radiographs, deep learning, super-resolution, domain adaptation, medical imaging, chest X-ray
\end{keywords}

\section{Introduction}
Coronary artery calcification (CAC) is a well-established marker of atherosclerotic burden and a strong predictor of cardiovascular events and all-cause mortality \cite{CAC-budoff}. Quantification using ECG-gated cardiac CT via the Agatston method remains the clinical gold standard for risk stratification \cite{agatston1990quantification, cutoff}. However, CT screening is costly, resource-intensive, and impractical for large-scale population use \cite{ct-screening}, motivating alternative modalities that are low-cost, scalable, and suitable for population-level risk assessment.

In contrast, chest X-rays (CXRs) are inexpensive and widely accessible, making them an attractive modality for opportunistic CAC risk assessment \cite{dancona2023deep}. Yet, CXRs have limited sensitivity for calcium depiction and lack standardised annotation protocols, leaving reliable ground truth labels scarce \cite{kamel2021prediction}. Existing CXR–CT datasets are typically assembled retrospectively, requiring patients to have undergone both imaging modalities within a clinically acceptable window (often $\leq$6 months). Such dual-modality acquisitions are uncommon in routine care and, when available, can introduce temporal misalignment and label noise due to disease progression between scans. This bottleneck constrains the development of deep learning methods, and prior studies \cite{kamel2021prediction, dancona2023deep, jeong2024radiomics} remain limited in scale and generalisability. 

In such data-scarce settings, an emerging strategy is to train models within surrogate domains: synthetic or simulated approximations of the target modality that preserve key imaging physics while offering abundant labels. In clinical domains, Digitally reconstructed radiographs (DRRs) are synthetic 2D projections of CT volumes that approximate real CXRs while being able to inherit precise CT-derived labels. Prior work in their generation has validated their clinical fidelity \cite{moore2011drr} suggesting that DRRs may serve as a scalable proxy training domain where methodological feasibility can be established before transferring to real CXRs. Although direct transfer to real CXRs is not evaluated here, recent work has shown DRR-trained models to generalize effectively in fluoroscopic and interventional settings, supporting the plausibility of similar transferability for CAC detection\cite{deepdrr}.

To our knowledge, this work provides the first systematic evaluation of DRRs as a surrogate training domain for CAC detection. We evaluate feasibility across model capacity, fidelity enhancement, preprocessing, and training strategies, and show that DRRs constitute a scalable and precisely labelled domain for developing models prior to transfer to real CXRs, laying the foundation for low-cost, population-scale cardiovascular risk screening using existing radiography infrastructure.

\section{Related Work}
\subsection{CT-based CAC detection}
ECG-gated cardiac CT is the clinical gold standard for CAC quantification via the Agatston method \cite{agatston1990quantification}, and numerous studies have leveraged CT directly for automated CAC detection and scoring using deep learning\cite{eng2021automated}. These approaches benefit from precise attenuation-based ground truth but remain limited by the cost, radiation dose, and infeasibility of CT screening at the population level \cite{ct-screening}.

\subsection{CXR-based CAC detection}
Given the wide availability of chest radiographs, several groups have explored their utility for CAC risk assessment. \cite{kamel2021prediction} trained an attention-augmented VGG16 on 1,689 CXRs paired with CTs, achieving an AUC of 0.73 for CAC prediction. \cite{dancona2023deep} demonstrated that deep learning on CXRs could refine pretest probability estimation in suspected angina patients, validated against invasive coronary angiography.\cite{jeong2024radiomics} proposed a radiomics-based approach requiring manual cardiac segmentation, reporting an AUC of 0.808 for detecting moderate-to-severe CAC ($>$100 Agatston units). Collectively, these approaches demonstrate the potential of CXRs for CAC screening, yet despite their promising results, progress remains constrained by the difficulty of assembling large paired CXR–CT datasets, the weak sensitivity of CXRs to calcification, and dependence on manual segmentation—factors that hinder scalability and reproducibility.

\subsection{Synthetic imaging with DRRs}
DRRs have been widely used in radiotherapy and orthopaedics to simulate radiographs from CT volumes, and prior work has validated their anatomical and clinical fidelity \cite{moore2011drr}. Recent studies have begun to explore DRRs as surrogate datasets for training deep learning models in scenarios where paired real-world imaging and labels are scarce \cite{deepdrr}. However, these efforts have primarily focused on registration or dose optimisation rather than disease detection. To date, no systematic evaluation has been conducted on their feasibility as a surrogate training domain for CAC detection.

\section{Methods}
\subsection{Dataset and Labels}
We use the publicly available Coronary Calcium and Chest CT (COCA) dataset \cite{coca2022}, which contains 790 ECG-gated cardiac CT scans paired with coronary artery calcium (CAC) segmentations in \texttt{xml} format. For each patient, a total Agatston score was computed by summing across per-artery calcium masks. Following clinical convention, we binarise labels at a threshold of 100: patients with scores $\leq$100 are classified as negative (no/mild CAC), and those $>$100 as positive (moderate/severe CAC). This threshold was not tuned, but chosen because it reflects the established clinical cutoff between non-actionable and clinically significant CAC \cite{cutoff}.  

To ensure geometric consistency, CT volumes were resampled to isotropic $\delta x$ mm spacing, where $\delta x$ was matched to the in-plane pixel spacing of the source DICOM. Patients with insufficient slice coverage ($s < 30$) were excluded for quality control, yielding 667 usable scans. Although gated CTs do not capture full thoracic coverage, they do represent the heart-centred regions most relevant for CAC detection. Comparable regions could be isolated in real CXRs with coarse heart localisation, suggesting that this dataset remains a reasonable surrogate for feasibility testing.

\subsection{Digitally Reconstructed Radiographs (DRRs)}
Synthetic radiographs were generated directly from CT volumes using the Siddon ray-tracing algorithm, implemented in the open-source \texttt{DiffDRR} framework \cite{diffdrr}. Siddon projection computes exact line integrals through the CT volume, avoiding interpolation artefacts and preserving small, high-density structures such as calcifications. For each scan, we simulated posterior–anterior (PA) and lateral (LA) projections under a fan-beam geometry with fixed source–detector distance (1085.6 mm) and detector width of 512 pixels at 1 mm spacing. Both PA and LA projections used identical source–detector parameters, with the LA view generated by a $90^{\circ}$ rotation around the cranio-caudal axis to approximate orthogonal orientation. DRRs were normalised to $[0,1]$ and resized to $512 \times 512$. This pipeline yields synthetic radiographs that approximate clinical CXRs while retaining precise CAC labels inherited from the source CTs.

\begin{figure}[!h]
\centering
\resizebox{\textwidth}{!}{
    \begin{tikzpicture}[
      font=\small,
      >=Latex,
      node distance=18mm and 18mm,
      arrow/.style={-Latex, line width=0.6pt},
      accent/.style={draw=black, fill=black!6, rounded corners, line width=0.6pt},
      img/.style={inner sep=0pt, outer sep=0pt},
      captext/.style={inner sep=1pt, text height=1.4ex, text depth=.3ex}
    ]

    \begin{scope}[shift={(0,0)}]
      \node[img, xshift=4mm, yshift=-4mm] (ct3) {\includegraphics[width=24mm]{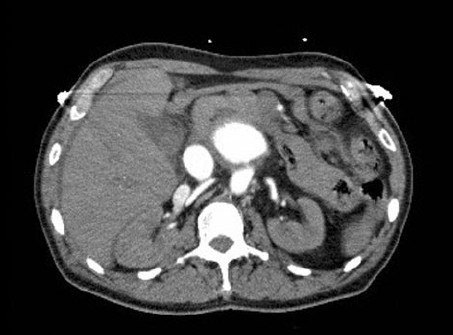}};
      \node[img, xshift=2mm, yshift=-2mm] (ct2) {\includegraphics[width=24mm]{ct_slice1.png}};
      \node[img]                           (ct1) {\includegraphics[width=24mm]{ct_slice1.png}};
      \begin{scope}[on background layer]
          \node[fit=(ct1)(ct2)(ct3), accent] (ctbox) {};
        \end{scope}
    \node[captext, below=2mm of ctbox] {CT Volume};
    \end{scope}
    
    \node[right=22mm of ctbox] (mid1) {};
    \draw[arrow] (ctbox.east) -- node[captext, above] {Projection} (mid1);
    \node[captext, left=3mm of mid1, yshift=-3mm] {\texttt{DiffDRR}};
    
    \node[img, right=-0.5mm of mid1] (drrimg) {\includegraphics[width=22mm]{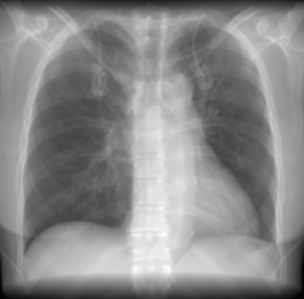}};
    \begin{scope}[on background layer]
        \node[fit=(drrimg), accent] (drrbox) {};
    \end{scope}
    \node[captext, below=2mm of drrbox]{Input DRR};
    
    \node[right=22mm of drrbox] (mid2) {};
    \draw[arrow] (drrbox.east) -- node[captext, above]{Preprocess}(mid2);
    
    \node[img, right=-0.5mm of mid2] (modelimg) {\includegraphics[width=30mm]{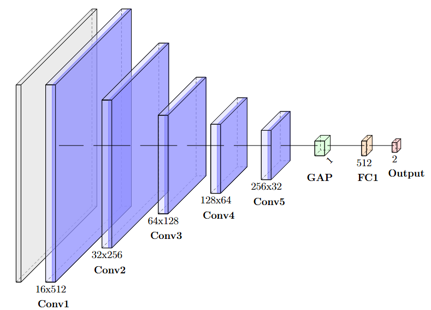}};
    \begin{scope}[on background layer]
        \node[fit=(modelimg), accent] (modelbox) {};
    \end{scope}
    \node[captext, below=2mm of modelbox]{ML Model};
    \node[accent, right=15mm of modelbox, minimum width=25mm, minimum height=12mm] (outputbox) {};
    \node[captext, font=\footnotesize] at (outputbox) {No CAC / CAC};
    
    \draw[arrow] (modelbox.east) -- (outputbox.west);
    \node[captext, below=8mm of outputbox]{Prediction};
    
    \end{tikzpicture}
    }
\caption{A CT volume (left) is projected into a DRR using Siddon’s algorithm (DiffDRR) and fed to a classifier to predict a binary CAC label (Agatston $>100$).}
\label{fig:min_pipeline}
\end{figure}
\vspace{-3mm}
\subsection{Image Enhancement Strategies}
To test whether DRRs provide sufficient fidelity for CAC detection—or whether additional enhancements are needed—we evaluate two complementary strategies: \textbf{(i)} pre-projection super-resolution and \textbf{(ii)} post-projection preprocessing.  

\subsubsection{Pre-projection super-resolution}
Native CTs often have anisotropic voxels, where coarse in-plane resolution, from a small number of axial slices, can obscure small calcifications once projected. To test whether resolution recovery enhances CAC depiction, we apply a $4\times$ SRResNet \cite{srgan} to sagittal slices prior to projection. The super-resolved slices were then reassembled into a volume, resampled isotropically, and projected identically to the native-resolution pipeline. This comparison probes whether fine detail restoration materially improves DRR-based detection.  

\subsubsection{Post-projection image preprocessing}
We further test whether DRR adjustments can aid detection by making CAC visually more salient. Three variants were compared:  
\textbf{(1) Original:} unaltered projection.  
\textbf{(2) CLAHE:} contrast-limited adaptive histogram equalisation to locally enhance soft tissue, implemented using an $8{\times}8$ tile grid and clip-limit of $2.0$, while the unsharp mask used a $5{\times}5$ Gaussian kernel ($\sigma{=}1.0$) with a gain of $1.5$. \textbf{(3) Calc-focused:} a composite filter designed to highlight calcifications and suppress irrelevant anatomy. Comprised of gamma correction $(\gamma=1.5)$, CLAHE and unsharp masking with a $5{\times}5$ Gaussian kernel ($\sigma{=}1.0$).
This axis examines whether heuristic contrast enhancement improves learnability, or whether native DRRs are sufficient.
\begin{figure}[!h]
    \centering
    \includegraphics[width=0.65\linewidth]{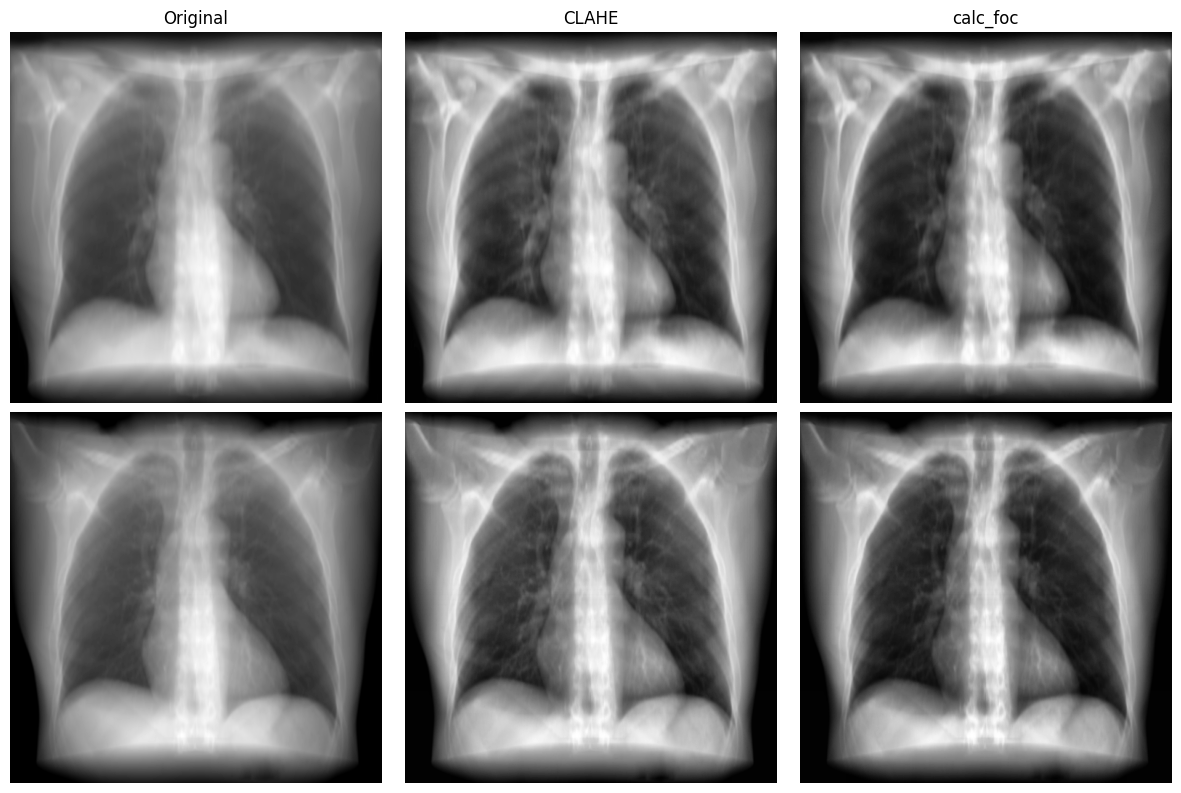}
    
    \vspace{1mm}
    \begin{minipage}{0.9\linewidth}
        \centering
        \hspace{0.0\linewidth}(a) Original \hspace{0.08\linewidth}(b) CLAHE \hspace{0.08\linewidth}(c) Calc-focused
    \end{minipage}

    \caption{Preprocessing for Native DRR (Top) and Super-resolved DRR (bottom)}
    \label{fig:preproc}
\end{figure}
\vspace{-4mm}
\subsection{Model Architectures}
To assess the robustness across model capacity and inductive bias, we evaluate three representative networks: 
\textbf{(i) CNN5\_GAP:} a lightweight custom CNN with five convolutional blocks, global average pooling, and a two-layer classifier. Each block comprised a $3{\times}3$ kernel with LeakyReLU activations and batch normalization, followed by dropout ($p{=}0.2$–$0.3$) in the classifier head for regularisation.
\textbf{(ii) ResNet18}\cite{resnet} a moderate-capacity residual network serving as a standard baseline. 
\textbf{(iii) DenseNet121}\cite{densenet-backbone} pretrained on CheXpert and augmented with a lightweight self-gating spatial attention mechanism derived from CBAM. More specifically, given feature maps $F \in \mathbb{R}^{B \times C \times H \times W}$, 
we compute a spatial attention mask 
$$M = \sigma\!\left(\frac{1}{C}\sum_{c=1}^{C} F_c \right)$$ 
where $\sigma(\cdot)$ denotes the sigmoid function. The attended feature map is then $\tilde{F} = F \odot M$, which reweights activations by their spatial salience prior to global pooling.

For multi-view experiments, we use dual-encoder variants incorporating postero-anterior (PA) and lateral (LA) projections. Fusion is tested at three levels: (i) early image concatenation, (ii) latent feature concatenation, and (iii) cross-attention. These designs examine whether DRRs encode complementary view information analogous to real CXR studies.
\begin{figure}[!h]
    \centering
    \includegraphics[width=0.4\linewidth]{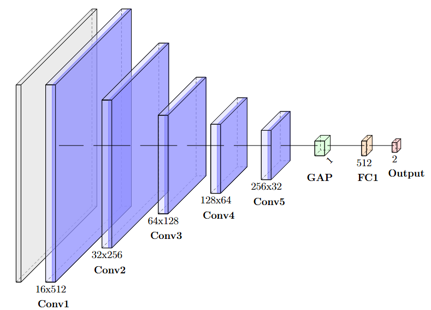}
    \caption{Architecture of the custom CNN5\_GAP model, consisting of five convolutional blocks and a 2-layer classifier.}
    \label{fig:cnn5_gap}
\end{figure}
\vspace{-4mm}

\subsection{Training \& Evaluation}
Models were trained with binary cross-entropy loss and label smoothing ($\varepsilon=0.1$) using Adam \cite{adam} (learning rate $10^{-4}$, weight decay $10^{-5}$, batch size 32). Augmentations consisted of random affine transformations: rotations ($\pm 5^{\circ}$), translations ($\pm 5\%$), scaling ($[0.9, 1.1]$) and shearing ($10^{\circ}$). Horizontal flips were excluded to preserve anatomical context. Early stopping was applied on validation AUC (being the criterion for model selection), with dropout/batch normalisation for regularisation. Multi-view fusion models used a curriculum schedule\cite{Curric} (extremes first, then borderline), intended as a heuristic to reduce instability from thresholding. 

Performance was evaluated with stratified 5-fold Cross Validation (CV) across 5 seeds (25 runs) with approximately 530 training and 130 validation DRRs. Metrics included AUC-ROC (primary), with accuracy, precision, recall, and F1 as secondary. All splits were patient-level and stratified by CAC label to avoid data leakage across folds. To reduce optimism bias from volatile early epochs, we discarded the first five epochs and averaged performance over the top five subsequent epochs. This conservative reporting may slightly underestimate peak metrics, but ensures robustness. Statistical significance was assessed using paired Wilcoxon signed-rank tests\cite{hollander2013} across validation folds. Results were considered significant at $p < 0.05$. 

All preprocessing, training, and evaluation code is implemented in PyTorch and will be made publicly available upon publication for reproducibility.
\section{Results}
We present results in a structured manner, beginning with single-view baselines and ablations on preprocessing and super-resolution, followed by multi-view fusion strategies, and finally training variants such as curriculum learning. Unless otherwise noted, all results are reported as mean validation AUC (± standard deviation) across 25 runs, with statistical significance assessed using the Wilcoxon signed-rank test.

\subsection{Baselines, Preprocessing, and Super-resolution}
To assess the impact of preprocessing and super-resolution (SR) on model performance, we evaluated three architectures (CNN5\_GAP, DenseNet121, and ResNet18) across three DRR preprocessing modes (Original, CLAHE, Calc-focused) under both native and super-resolved inputs (Table \ref{tab:preproc_sr}). For CNN5\_GAP, mean AUCs improved by $+0.015$  ($0.720\rightarrow0.735$) from native-original to SR-calc-focused, with SR yielding significant gains under calc-focused and CLAHE preprocessing ($p<0.05$). DenseNet121 performance degraded $(0.718 \rightarrow 0.688;\,\,p < 0.05)$ under contrast-enhancing preprocessing, with SR offering only minor recovery. ResNet18 remained stable across most modes, with no significant differences observed. 

This divergence highlights that pretrained backbones, while strong under natural-image statistics, may be brittle to synthetic-domain contrast shifts. Thus, given its consistent and competitive performance, CNN5\_GAP was selected as the primary backbone for subsequent experiments, suggesting that compact task-specific CNNs trained from scratch may generalise better within synthetic domains where texture and contrast statistics could differ to real-world pretraining distributions.
\begin{table}[!h]
\centering
\caption{Effect of preprocessing and super-resolution on validation AUC for CNN5\_GAP, augmented DenseNet121 and ResNet18.\protect\footnotemark}
\label{tab:preproc_sr}
\resizebox{\linewidth}{!}{
\begin{tabular}{lcccccc}
\toprule
\multirow{2}{*}{Preprocessing} & \multicolumn{2}{c}{CNN5\_GAP} & \multicolumn{2}{c}{DenseNet121} & \multicolumn{2}{c}{ResNet18} \\
 & Native & SR & Native & SR & Native & SR \\
\midrule
Original & 0.720 ± 0.048 & 0.727 ± 0.050 & \textbf{0.718 ± 0.031$^\dagger$} & 0.708 ± 0.026 & 0.714 ± 0.031 & 0.717 ± 0.031 \\
CLAHE    & 0.728 ± 0.049 & \textbf{0.733 ± 0.043$^*$} & 0.688 ± 0.053 & 0.695 ± 0.028 & 0.716 ± 0.068 & 0.702 ± 0.022 \\
Calc-foc & 0.729 ± 0.053 & \textbf{0.735 ± 0.043$^*$} & 0.683 ± 0.058 & 0.688 ± 0.042 & 0.722 ± 0.070 & 0.723 ± 0.032 \\
\bottomrule
\end{tabular}}
\end{table}
\footnotetext{$\dagger$ $p < 0.05$ vs CLAHE, Calc\_foc $*$ $p < 0.05$ vs Original}
\vspace{-5mm}
\subsection{Fusion \& Training Strategies}
We next evaluated whether combining posterior–anterior (PA) and lateral (LA) projections could improve performance over best performing single-view (PA) baseline of $0.735 \pm 0.043$. Three fusion strategies were tested: early fusion (image concatenation), intermediary fusion (latent feature concatenation), and cross-attention fusion. Each was implemented with either shared or unshared encoders and a learnable scalar-gated mechanism to weigh each view's contribution.

\begin{table}[!h]
\centering
\caption{Performance of PA+LA fusion strategies, varying by level of interaction (early, intermediary, or attention) and whether encoders are shared or unshared.}
\label{tab:fusion_results}
\resizebox{0.7\linewidth}{!}{
\begin{tabular}{lcc}
\toprule
Fusion Strategy & Shared Encoder & Unshared Encoder \\
\midrule
PA-Only (Baseline)         & $0.735 \pm 0.043$ \\
Early Fusion             & 0.702 ± 0.044 & -- \\
Intermediate Fusion      & 0.736 ± 0.059 & 0.739 ± 0.065 \\
Cross-Attention Fusion   & \textbf{0.740 ± 0.043} & 0.729 ± 0.045 \\
\bottomrule
\end{tabular}}
\end{table}
As shown in Table \ref{tab:fusion_results}, early fusion was significantly worse ($p < 0.05$) relative to the PA-only baseline ($0.735 \pm 0.043$). Intermediary fusion achieved mean AUCs of 0.736 (shared encoders) and 0.739 (unshared), while cross-attention fusion reached 0.740 (shared) and 0.729 (unshared). Differences among fusion variants were within one standard deviation, and none reached statistical significance \textit{relative} to the PA-only baseline. These results suggest that PA projections carry the dominant discriminative signal and that complementary anatomical information from orthogonal views may already be implicitly captured under limited-data conditions, rather than indicating methodological failure. Consistent with Appendix \ref{appendix:C}, LA-only performance (0.695) trailed PA substantiating this claim.

We also assessed the effect of alternative training strategies on the best-performing fusion model (cross-attention/shared). Specifically, we evaluated curriculum learning, SimCLR-based self-supervised pretraining\cite{simclr}, and their combination.

\begin{table}[!h]
\caption{Final fusion model performance across training strategies. SimCLR and curriculum learning offer complementary benefits.}
\centering
\begin{tabular}{lc}
\toprule
Training Strategy & Mean AUC \\
\midrule
Fusion (standard)        & 0.740 $\pm$ 0.043 \\
Curriculum Learning      & 0.750 $\pm$ 0.046 \\
SimCLR Pretraining       & 0.742 $\pm$ 0.045 \\
SimCLR + Curriculum      & \textbf{0.754 $\pm$ 0.055} \\
\bottomrule
\end{tabular}
\label{tab:simclr_curric}
\end{table}

Table \ref{tab:simclr_curric} shows that curriculum learning yielded a higher mean AUC from 0.740 to 0.750. SimCLR pretraining yielded 0.742, close to the baseline. SimCLR combined with curriculum achieved the highest observed mean AUC (0.754), however this was comparable within variance. The consistent direction of improvement across folds suggests potential value, but larger datasets are required for confirmation (See Appendix \ref{appendix:C}, Table \ref{tab:wilcoxon_results} for pairwise Wilcoxon statistics). These results motivate examining how view complementarity and training stability interact under limited data, as discussed next
\section{Discussion}
Our experiments evaluated the feasibility of using DRRs as a proxy domain for CAC detection, focusing on: \textbf{(i)} fidelity enhancement via super-resolution, \textbf{(ii)} signal optimisation through preprocessing, and \textbf{(iii)} fusion of complementary projections.

\subsection{Super-resolution and Preprocessing}

Calcified lesions are high-frequency features and their visibility is diminished when CT volumes are reconstructed at low axial resolutions. On its own, SR modestly improved fidelity, yielding consistent but non-significant gains in CNN5 GAP ($+0.007;p=0.107$). When paired with preprocessing, SR-enabled pipelines achieved improvements over their native baselines ($p < 0.05$), with Calc\_foc + SR yielding the strongest single-view performance (0.735). This suggests that SR may play an enabling role: by partially restoring high-frequency structure, it could provide the substrate that preprocessing methods can then amplify into a more discriminative signal.

Preprocessing alone showed model-dependent behaviour. CNN5\_GAP adapted well, whereas DenseNet121 degraded significantly ($p < 0.05$). This highlights a trade-off: pretrained backbones encode strong priors but are brittle to contrast shifts, while lightweight task-specific models are more flexible and can exploit enhanced contrast once fidelity is recovered. However, real CXRs are already acquired at high resolutions; SR's benefit is unique to CT-derived projections. In contrast,  preprocessing strategies transfer rather trivially. In clinical CXRs, where acquisition noise and patient variability are greater, local contrast enhancement or normalisation may provide more value by homogenising inputs while highlighting subtle calcifications. 

Taken together, these results suggest that fidelity restoration and contrast enhancement jointly determine the effective signal-to-noise ratio for CAC depiction in DRRs 

\subsection{Fusion \& Training Strategies}
Given calcifications may be obfuscated in a single projection, we evaluated fusion of PA and LA DRRs. Early image-level fusion performed significantly worse ($p < 0.05$), suggesting destructive interference across modalities. Intermediary feature fusion and cross-attention achieved the best observed gains, though improvements did not reach statistical significance, likely reflecting fold-level variance under limited data. Notably, LA-only classification underperformed (AUC 0.695, see Table \ref{tab:la_performance}), indicating that PA projections provide most discriminative information, while the marginal gain from incorporating LA views may be limited under current dataset size and coverage.
Curriculum learning stabilised training (+0.010 AUC), SimCLR offered marginal gains, and their combination reached the highest mean AUC (0.754), consistent across folds but not statistically significant. Their complementary effects likely stem from stabilising optimisation (curriculum) and improving representation initialisation (SimCLR). Ultimately, these observations highlight the need for larger, more diverse datasets to fully exploit multi-view complementarity and self-supervised pretraining. 

Overall, SR enhances DRR fidelity for lightweight CNNs, preprocessing benefits depend on architecture, and fusion trends are promising but data-limited. DRRs reproduce key behaviours seen in real CXR studies while providing label-rich CT-derived data, substantiating their value as a proxy domain and motivating future work in domain-adaptation.

\subsection{Comparison to Prior Work}
Our best-performing configuration (mean AUC = 0.754) is comparable to, and in some cases slightly exceeds, results from prior CXR-based CAC detection studies. \cite{kamel2021prediction} reported an AUC of 0.73 using an attention-augmented VGG16 trained on 1,689 paired CXRs and \cite{dancona2023deep} achieved $\approx$0.71 AUC when predicting significant coronary artery disease from radiographs. In contrast to \cite{jeong2024radiomics} whose radiomics-based approach required manual cardiac segmentation, our pipeline is fully end-to-end and does not rely on handcrafted features or manual delineation. Additionally, our approach offers two further methodological advantages: \textbf{(i)} precise CT-derived labels free from annotation noise and \textbf{(ii)} scalability unconstrained by the scarcity of dual-modality data.

Our best-performing configuration (mean AUC = 0.754 ± 0.055) is comparable to, and in some cases slightly exceeds, previously reported results from real CXR–based CAC detection studies. \cite{kamel2021prediction} achieved an AUC of 0.73 using an attention-augmented VGG16 trained on 1,689 paired CXRs and CTs, while \cite{dancona2023deep} reported approximately 0.71 for detecting significant coronary artery disease from radiographs. \cite{jeong2024radiomics} reached an AUC of 0.808 using a radiomics-based pipeline requiring manual cardiac segmentation.

In contrast, our pipeline is \textbf{fully automated and end-to-end}, requiring no manual segmentation or handcrafted features, and achieves comparable discriminative performance despite being trained on a smaller dataset of synthetic DRRs. Crucially, our approach inherits precise, noise-free labels directly from CT-derived Agatston scores, avoiding the temporal misalignment and label uncertainty that affect retrospective CXR–CT pairing studies.

Moreover, the proposed DRR-based framework is \textbf{intrinsically scalable and reproducible}. Once a labelled CT repository is available, thousands of perfectly aligned DRRs can be generated automatically under controlled imaging geometry. This removes the bottleneck of acquiring dual-modality CXR–CT pairs and establishes a standardized, label-rich training domain that can be shared and benchmarked across studies.

These characteristics enable our synthetic DRR pipeline to achieve near state-of-the-art AUC performance within a controlled and reproducible framework, offering a pragmatic foundation for developing and validating CAC detectors prior to transfer to real CXRs.

\section{Limitations}
Although results are promising, several limitations should be acknowledged. Since Agatston scores are aggregated across coronary arteries, they provide only image-level supervision. While appropriate for weakly-supervised feasibility, future work should explore region-localized DRR labels derived from per-artery calcium masks to better capture spatial heterogeneity. Further, the dataset remains small by deep-learning standards with validation folds containing about 130 DRRs, contributing to fold-level variance and modest effect sizes. More extensive cohorts and higher-fold cross-validation would clarify the validity of observed trends. Importantly, the synthetic domain may not fully replicate the characteristics of clinical CXRs. Differences in noise statistics, detector response, and cardiac field-of-view mean that models trained purely on DRRs may not transfer directly to real radiographs. Thus, the present study demonstrates methodological feasibility rather than clinical readiness. Future work should therefore prioritise domain-adaptation strategies and real-CXR validation to establish relevance.

\section{Conclusion}
This study demonstrates the feasibility of using CT-derived DRRs as a proxy training domain for DL–based CAC detection, offering a scalable alternative in the face of scarce labelled CXR datasets. We showed that lightweight architectures trained from scratch can outperform larger pretrained models, and that pairing SR with contrast-enhancing preprocessing yields statistically significant performance gains, supporting the value of fidelity recovery in DRRs. Fusion of orthogonal projections produced the best observed performance despite not reaching significance under data-limited evaluation. Structured training strategies such as curriculum learning improved robustness under label noise, with SimCLR pretraining offering only marginal benefit. Beyond these methodological insights, our findings underscore the broader potential of DRRs as a stable, label-rich surrogate for CXRs. By establishing a viable synthetic environment, this framework lowers the barrier for reproducible CAC detection research and could serve as a pretraining resource for future CXR-based screening pipelines. With further work in domain adaptation, dataset expansion, and multimodal integration, DRR-trained models could in principle form the basis of clinically deployable screening tools for population-level cardiovascular risk assessment.

\clearpage  
\midlacknowledgments{The authors are grateful for the support by the 2024 (Cardiac, Vascular, and Metabolic Medicine) CVMM THEME COLLABORATIVE GRANT SCHEME.}

\newpage
\appendix
\section{Statistical Testing Methods}
For each paired comparison (same folds/seeds), we apply the Wilcoxon signed-rank test on per-run AUC values. Using 5 randomised seeds across a 5-fold CV, this yields $n=25$ runs.  We report the test statistic and p-value. Significant results are marked as bold at $p<0.05$ with a $*$.

\section{Gated Dataset}
\label{appendix:A}
\begin{table}[!h]
\centering
\caption{Gated dataset – Wilcoxon signed-rank test within CNN5\_GAP across super-resolution and preprocessing modes. Statistical significance at $p<0.05$ is marked with *.}
\begin{tabular}{lcccc}
\hline
Comparison & Stat & $p$-value & Mean AUC A & Mean AUC B \\
\hline
Native: orig vs calc\_foc & 97.0 & 0.078 & 0.720 & 0.729 \\
Native: orig vs clahe & 104.0 & 0.120 & 0.720 & 0.728 \\
Native: calc\_foc vs clahe & 151.0 & 0.771 & 0.729 & 0.728 \\
SR: orig vs calc\_foc & 66.0 & \sig{8.07e-03} & 0.727 & 0.735 \\
SR: orig vs clahe & 89.0 & \sig{0.048} & 0.727 & 0.733 \\
SR: calc\_foc vs clahe & 123.0 & 0.300 & 0.735 & 0.733 \\
Native vs SR (orig) & 129.0 & 0.381 & 0.720 & 0.727 \\
Native vs SR (calc\_foc) & 102.0 & 0.107 & 0.729 & 0.735 \\
Native vs SR (clahe) & 91.0 & 0.055 & 0.728 & 0.733 \\
\hline
\end{tabular}
\end{table}

\begin{table}[!h]
\centering
\caption{Gated dataset – Wilcoxon signed-rank test within DenseNet121 across super-resolution and preprocessing modes. Statistical significance at $p<0.05$ is marked with *.}
\begin{tabular}{lcccc}
\hline
Comparison & Stat & $p$-value & Mean AUC A & Mean AUC B \\
\hline
Native: orig vs calc\_foc & 83.0 & \sig{0.032} & 0.718 & 0.683 \\
Native: orig vs clahe & 81.0 & \sig{0.028} & 0.718 & 0.688 \\
Native: calc\_foc vs clahe & 122.0 & 0.276 & 0.683 & 0.688 \\
SR: orig vs calc\_foc & 66.5 & \sig{0.010} & 0.708 & 0.688 \\
SR: orig vs clahe & 80.0 & \sig{0.026} & 0.708 & 0.695 \\
SR: calc\_foc vs clahe & 128.0 & 0.367 & 0.688 & 0.695 \\
Native vs SR (orig) & 105.0 & 0.122 & 0.718 & 0.708 \\
Native vs SR (calc\_foc) & 150.0 & 0.751 & 0.683 & 0.688 \\
Native vs SR (clahe) & 129.0 & 0.381 & 0.688 & 0.695 \\
\hline
\end{tabular}
\end{table}

\newpage
\section{Fusion Analysis}
\label{appendix:C}
\subsection{Lateral-only Performance}
Here we assess the strength of the calcium signal in Lateral (LA) projections. Significance was assessed with Wilcoxon signed-rank tests at $p < 0.05$.
\begin{table}[!h]
\centering
\caption{Wilcoxon signed-rank test results for LA-only performance across preprocessing modes using CNN5\_GAP. 
Statistical significance at $p<0.05$ is marked with *.}
\label{tab:la_performance}
\begin{tabular}{lcccc}
\toprule
Comparison & Stat & $p$-value & Mean AUC A & Mean AUC B \\
\midrule
LA Native: orig vs calc\_foc & 37.0 & \sig{3.29e-04} & 0.672 & 0.695 \\
LA Native: orig vs clahe & 57.0 & \sig{0.003} & 0.672 & 0.691 \\
LA Native: calc\_foc vs clahe & 120.5 & 0.258 & 0.695 & 0.691 \\
\bottomrule
\end{tabular}
\end{table}

\subsection{Fusion and Training Strategy Comparisons}
Pairwise Wilcoxon signed-rank test results comparing fusion baselines, curriculum learning, and SimCLR variants. Significant results at $p < 0.05$ are marked with $*$.
\begin{table}[!h]
\centering
\caption{Pairwise Wilcoxon signed-rank test results comparing fusion baselines, curriculum learning, and SimCLR variants. Statistical significance at $p<0.05$ is marked with *.}
\label{tab:wilcoxon_results}
\begin{tabular}{lcccc}
\toprule
Comparison & Stat & $p$-value & Mean AUC A & Mean AUC B \\
\midrule
Early vs Int. Fusion (Shared) & 11.0 & \sig{4.6e-05} & 0.703 & 0.736 \\
Early vs Int. Fusion (Unshared) & 11.0 & \sig{4.6e-05} & 0.703 & 0.740 \\
Early vs Cross-Attention (Shared) & 154.0 & 0.833 & 0.739 & 0.740 \\
\midrule
Standard vs Curriculum & 132.0 & 0.426 & 0.740 & 0.750 \\
Standard vs SimCLR & 152.0 & 0.791 & 0.740 & 0.742 \\
Standard vs Curriculum + SimCLR & 128.0 & 0.367 & 0.740 & 0.754 \\
\bottomrule
\end{tabular}
\end{table}

\end{document}